\documentclass[journal]{IEEEtran}
\usepackage[T1]{fontenc}
\usepackage{graphicx}
\usepackage{times}
\usepackage{helvet}
\usepackage{courier}
\usepackage{amsmath}
\usepackage{algorithm}
\usepackage{algorithmic}
\usepackage{csquotes} 
\usepackage{color}
\usepackage{paralist}
\usepackage{amssymb}
\usepackage{indentfirst}
\usepackage{subfigure}
\usepackage{float}
\usepackage{multirow}
\usepackage{cite}
\usepackage{mathrsfs}

\usepackage{mathrsfs} 
\usepackage{amsfonts}
\usepackage{todonotes}
\usepackage{pgfplots} 
\usepackage{epstopdf}
\usepackage{epsfig}
\pgfplotsset{compat=newest}
\usepackage{color}
\definecolor{forestgreen}{RGB}{0,139,69}

\usepackage{xcolor}
\definecolor{citecolor}{HTML}{0071bc}
\usepackage[colorlinks, linkcolor=red,  anchorcolor=blue, citecolor=citecolor]{hyperref} 

\usepackage{xcolor}
\definecolor{SeaGreen4}{RGB}{0,205,102} 
\definecolor{SlateBlue}{RGB}{106,90,205} 
\definecolor{DarkRed}{RGB}{178,34,34} 
	
\usepackage[switch]{lineno}

\usepackage{textcomp,booktabs}
\usepackage{amssymb}
\usepackage{pifont}

\usepackage{makecell}

\usepackage{colortbl}
\definecolor{mygray}{gray}{.9}
\definecolor{mypink}{rgb}{.99,.91,.95}
\definecolor{mycyan}{cmyk}{.3,0,0,0}

\begin{document}

\title{ 
RGB-Event HyperGraph Prompt for Kilometer Marker Recognition based on Pre-trained Foundation Models  
}

\author{Xiaoyu Xian, Shiao Wang, Xiao Wang, \emph{Member, IEEE}, Daxin Tian*, Yan Tian

\thanks{ $\bullet$ Xiaoyu Xian is with School of Transportation Science and Engineering, Beihang University, Beijing 100191, China; CRRC Academy Co., Ltd, Beijing, 100070, China (email: xxy@crrc.tech)} 

\thanks{ $\bullet$ Shiao Wang and Xiao Wang are with the School of Computer Science and Technology, Anhui University, Hefei 230601, China. (email: \{xiaowang\}@ahu.edu.cn, wsa1943230570@126.com)} 

\thanks{ $\bullet$ Daxin Tian is with School of Transportation Science and Engineering, Beihang University, Beijing 100191, China (email: dtian@buaa.edu.cn)} 

\thanks{ $\bullet$ Yan Tian is with CRRC Academy Co., Ltd, Beijing, 100070, China (email: ty@crrc.tech)}

\thanks{ * Corresponding Author: Daxin Tian} 
}

\markboth{ IEEE Transactions on Cognitive and Developmental Systems (IEEE TCDS) , 2026 } 
{Shell \MakeLowercase{\textit{et al.}}: Bare Demo of IEEEtran.cls for IEEE Journals}

\maketitle

\begin{abstract}
Metro trains often operate in highly complex environments, characterized by illumination variations, high-speed motion, and adverse weather conditions. These factors pose significant challenges for visual perception systems, especially those relying solely on conventional RGB cameras. To tackle these difficulties, we explore the integration of event cameras into the perception system, leveraging their advantages in low-light conditions, high-speed scenarios, and low power consumption. Specifically, we focus on Kilometer Marker Recognition (KMR), a critical task for autonomous metro localization under GNSS-denied conditions. In this context, we propose a robust baseline method based on a pre-trained RGB OCR foundation model, enhanced through multi-modal adaptation. Furthermore, we construct the first large-scale RGB-Event dataset, EvMetro5K, containing 5,599 pairs of synchronized RGB-Event samples, split into 4,479 training and 1,120 testing samples. Extensive experiments on EvMetro5K and other widely used benchmarks demonstrate the effectiveness of our approach for KMR.
Both the dataset and source code will be released on \url{https://github.com/Event-AHU/EvMetro5K_benchmark}
\end{abstract}


\IEEEpeerreviewmaketitle

\section{Introduction} 

\IEEEPARstart{M}{etro} systems play a crucial role in urban transportation, serving as the backbone of smart cities and intelligent transportation infrastructures. Their operation faces numerous challenges, particularly in achieving accurate train positioning and maintaining precise speed control, both of which are critical for ensuring safety, punctuality, and overall operational efficiency. These challenges are further compounded by dynamic and often harsh environmental conditions, such as dim lighting in underground tunnels, excessive sunlight exposure in above-ground sections, and interference from adverse weather. Under such circumstances, conventional RGB cameras frequently struggle to provide reliable perception, especially during high-speed train motion and in low-illumination scenarios, highlighting the need for more robust sensing solutions capable of handling these extreme operating conditions. 

To address the aforementioned issues, existing works~\cite{he2022masked, radford2021learning} have explored pre-training on large-scale RGB datasets to achieve satisfactory performance. For instance, the masked autoencoder (MAE) strategy~\cite{he2022masked}, built upon the Vision Transformer (ViT)~\cite{dosovitskiy2020image}, enhances the model’s visual perception ability by masking and reconstructing partial image patches. Meanwhile, CLIP~\cite{radford2021learning} learns semantic alignment between images and texts, thereby enabling strong cross-modal understanding and matching capabilities. However, due to the inherent limitations of RGB data under extreme conditions, it remains challenging to capture sufficient target details for effective perception.


Recently, event cameras have drawn more and more attention due to their advantages in high dynamic range, low energy consumption, and high temporal resolution~\cite{gallego2020eventsurvey}. The imaging principle of event cameras differs from that of conventional visible light cameras. Event cameras operate by measuring the brightness variations of pixels in a scene and emit asynchronous event signals when the change exceeds a specific threshold, outputting event points with polarity (+1, -1). These cameras have a higher dynamic range of approximately 120 dB, far surpassing the 60 dB range of the widely used RGB cameras, making them perform significantly better in overexposure and low-light scenarios. Additionally, due to their sparsity in the spatial and density in the temporal, event cameras excel at capturing fast-moving objects.

Combining RGB cameras with event cameras to achieve robust and reliable visual perception has gradually become a hot research trend, for example, RGB-Event based visual tracking~\cite{tang2022revisiting, wang2024event, huang2024mamba}, object detection~\cite{tomy2022fusing, zhou2022rgb}, pattern recognition~\cite{wang2024hardvs, wang2025rgb}, sign language translation~\cite{wang2025sign}. It is also widely used in many low-level image/video processing tasks, such as denoising~\cite{ding2023mlb}, super resolution~\cite{wang2020eventsr}. Specifically, Huang et al.~\cite{huang2024mamba} achieve effective visual object tracking by integrating event cameras with RGB cameras using a linear-complexity Mamba network. Wang et al.~\cite{wang2025rgb} propose a pedestrian attribute recognition task based on RGB-event modalities, and introduce the first large-scale multi-modal pedestrian attribute recognition dataset EventPAR, as well as an innovative RWKV fusion framework.

\begin{figure*}
\centering
\includegraphics[width=0.95\linewidth]{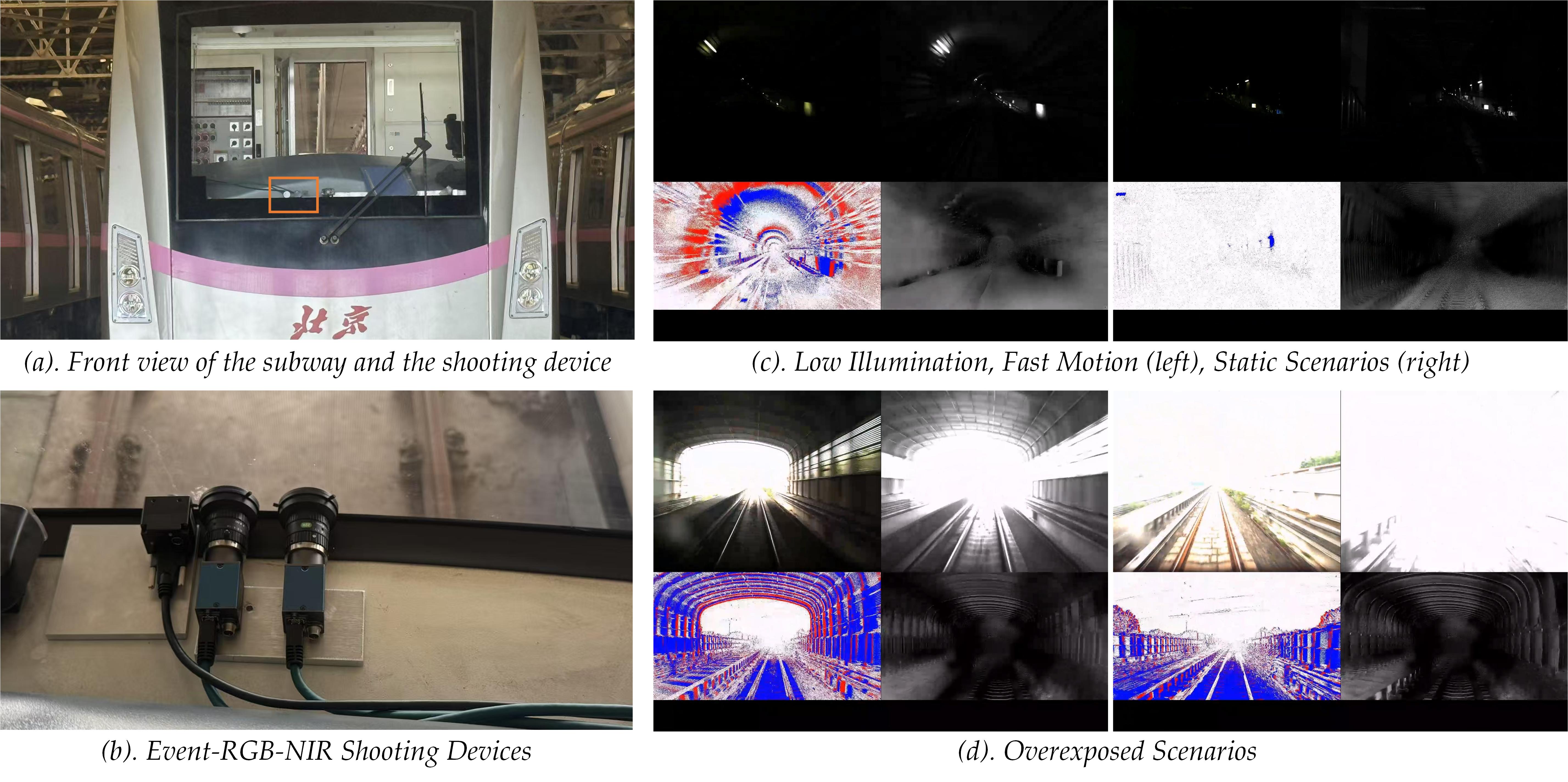}
\caption{The multi-modal imaging device proposed in this paper and typical metro perception scenarios. Specifically, sub-figures (c) and (d) show the imaging results of the metro under low-illumination conditions in high-speed motion/static scenes and overexposed scenes, respectively. In each sub-figure (c, d), the four images arranged clockwise are: the RGB image, the NIR image, the stacked event stream rendered as a red-blue map, and the reconstructed grayscale image from the event stream.}
\label{fig:firstIMG}
\end{figure*}

The visual localization method based on recognizing metro kilometer markers is an effective approach to achieve precise positioning of metros under GNSS-denied conditions. However, its performance often limited by the complex operating environment of metros, such as high-speed motion, dim tunnel lighting, and extreme weather, which severely degrades the recognition accuracy of RGB single-modality vision.
In this work, we resort to the RGB-Event multi-modal fusion for the Kilometer Marker Recognition (KMR), which fully leverages the advantages of event-based sensors in low-light and high-speed motion scenarios. 

To this end, we propose a robust baseline approach for KMR, referred to as HGP-KMR. Given the RGB and Event streams, we first reconstruct the grayscale image from the event stream using an events-to-grayscale image reconstruction algorithm~\cite{rebecq2019high}. Then, we embed the RGB frames into the vision tokens and extract their features using the vision Transformer layer. For the grayscale image, we first concatenate the RGB and event tokens into unified feature representations and construct a multi-modal hypergraph. Two hypergraph convolutional layers are adopted to encode the hypergraph and add the enhanced features into the RGB branch for multi-modal fusion. We integrate our modules into the PARseq~\cite{bautista2022PARseq} for accurate RGB-Event Kilometer Marker Recognition. An overview of our proposed framework can be found in Fig.~\ref{framework}. 
{Different from existing hypergraph-based fusion methods~\cite{wang2025evraindrop, gao2024hypergraph}, our work employs a hypergraph network to capture high-order interactions between RGB and event modalities, enabling richer cross-modal relationship learning. The hypergraph prompting strategy, in which multimodal hypergraph features guide and modulate the RGB branch, enhances robustness under noisy or degraded conditions.}

Beyond this, this paper formally proposes, for the first time, the use of RGB-Event cameras to recognize metro mileage. Specifically, we first set up a multi-modal imaging system, consisting of an RGB camera (MER2-134-90GC-P), a near-infrared camera (MER2-134-90GM-P), and a high-definition event camera (Prophesee EVK4), as shown in Fig.~\ref{fig:firstIMG}. Based on this multi-modal equipment, we collected more than 20 hours of multi-modal videos in the metro scenarios, capturing diverse conditions including different weathers (sunny, cloudy, rainy), times of day, and lighting conditions (daytime and nighttime scenarios), and train speeds. From these videos, we extracted 5599 effective mileage samples, which were manually annotated and verified, resulting in the construction of the EvMetro5K dataset. Using this dataset, we extended conventional unimodal character recognition algorithms to multi-modal versions through feature-level fusion, and retrained and evaluated these models to establish a comprehensive benchmark. This dataset and benchmark lay a solid foundation for future research on Kilometer Marker Recognition.

To sum up, the main contributions of this paper can be summarized as the following three aspects: 


1). We have developed the first RGB-Event perception imaging system in the railway transportation domain and acquired a large-scale multi-modal dataset, termed EvMetro5K. It can effectively support vision-based perception tasks, such as Kilometer Marker Recognition and dynamic scene reconstruction. 

2). We propose a novel RGB-Event HyperGraph Prompt for Kilometer Marker Recognition based on pre-trained foundation models, termed HGP-KMR. This method fully associates high-order information from both modalities through a multi-modal hypergraph prompt, achieving more robust and accurate recognition. 

3). Extensive experiments on our newly proposed EvMetro5K dataset, WordArt*, and IC15* datasets fully demonstrate the effectiveness of our proposed framework.

\textit{The following of this paper is organized as follows:} 
In section~\ref{sec::relatedWorks}, we give an introduction to the related works, including scene text recognition, RGB-Event fusion, and large foundation models. After that, we formally propose our framework in section~\ref{sec::method} and the newly built benchmark dataset in section~\ref{sec::benchmark}. The experiments are conducted in section~\ref{sec::experiments}. Finally, we conclude this paper in section~\ref{sec::conclusion}.


\section{Related Works} \label{sec::relatedWorks} 
 
\subsection{Scene Text Recognition}  

Scene text recognition (STR)~\cite{wang2011end,shi2016end,liao2019scene,han2024spotlight} inherently combines vision and language understanding. Early research mainly focused on either visual feature extraction or linguistic modeling, whereas recent approaches emphasize their integration for robustness under diverse conditions.  
E2STR~\cite{zhao2024E2STR} enhances adaptability by introducing context-rich text sequences and a context training strategy.  
CCD~\cite{Guan2023CCD} leverages a self-supervised segmentation module and character-to-character distillation, while SIGA~\cite{guan2023SIGA} further refines segmentation through implicit attention alignment.  
CDistNet~\cite{zheng2024cdistnet} incorporates visual and semantic positional embeddings into its transformer-based design to handle irregular text layouts.  
In parallel, iterative error correction strategies have been introduced via language models.    
VOLTER~\cite{li2024volter}, BUSNet~\cite{Wei2024BUSNet}, MATRNet~\cite{na2022matrn}, LevOCR~\cite{da2022levocr}, and ABINet~\cite{fang2021ABINet} exemplify this trend by refining recognition through feedback loops.  
Recently, large language model (LLM)-based STR has emerged, exploiting generative and contextual reasoning to unify visual-linguistic understanding.  
Representative works include TextMonkey~\cite{liu2024textmonkey}, DocPedia~\cite{feng2023docpedia}, Vary~\cite{wei2025vary}, mPLUG-DocOwl 1.5~\cite{hu2024mplug}, and OCR2.0~\cite{wei2024OCR2.0}.  
Despite their progress, these LLM-based models remain vulnerable under extreme conditions such as low illumination, blur, and noise.  

To overcome the limitations of RGB-only recognition, recent research explores event camera data for STR. EventSTR~\cite{wang2025eventstr} pioneers this direction, leveraging the high dynamic range and low-latency properties of event streams to improve robustness in adverse environments. Furthermore, ESTR-CoT~\cite{wang2025estrCoT} introduces chain-of-thought style reasoning with event representations, enhancing recognition accuracy through structured linguistic refinement. 
Generally speaking, existing works primarily focus on using RGB or event cameras to address text recognition tasks, but few explore the integration of multi-modal data for recognition, which limits their perceptual capability in complex scenarios.

\subsection{RGB-Event Fusion} 
Event cameras possess high temporal resolution and inherent robustness to motion blur, making them valuable complements to RGB data in various vision tasks to enhance overall performance. 
{Existing RGB-Event fusion methods can be broadly categorized into implicit modeling and explicit modeling approaches. For instance, Zhang et al.~\cite{zhang2022data} implicitly associate event streams with intensity frames under different baseline conditions, whereas Chen et al.~\cite{chen2025evhdr} achieve cross-modal fusion by explicitly constructing novel relationships between event streams and LDR images.
Leveraging the unique advantages of event cameras, RGB-Event fusion has demonstrated significant effectiveness across multiple applications. Specifically, their high dynamic range enables accurate HDR reconstruction~\cite{yang2023learning, chen2025evhdr} and robust low-light enhancement~\cite{chen2024event, liang2024towards}. Moreover, thanks to their superior temporal resolution, event cameras facilitate precise dynamic scene reconstruction~\cite{lu2025edygs, liu2025nemf} and effective video super-resolution~\cite{yan2025evstvsr}.} 
In addition, in image deblurring, Color4E~\cite{ma2024color4e} and CFFNet~\cite{li2024coarse} integrate event information with RGB images to recover sharp details under motion blur or low-light conditions. For multi-modal perception, UTNet~\cite{guo2025utnet} fuses events and RGB frames, effectively improving detection performance of transparent underwater organisms and enabling more accurate and efficient scene understanding. In 3D reconstruction and neural rendering, E2NeRF~\cite{qi2023e2nerf} leverages event data to compensate for motion blur and generate high-quality volumetric representations. In object detection, SFDNet~\cite{fan2025efficient} combines RGB and event data to improve robustness in dynamic scenes. Inspired by these works, this paper attempts to leverage RGB-Event data and achieve high-performance Kilometer Marker Recognition in challenging scenarios through hypergraph prompt learning. 

\begin{figure*}
\centering
\includegraphics[width=\textwidth]{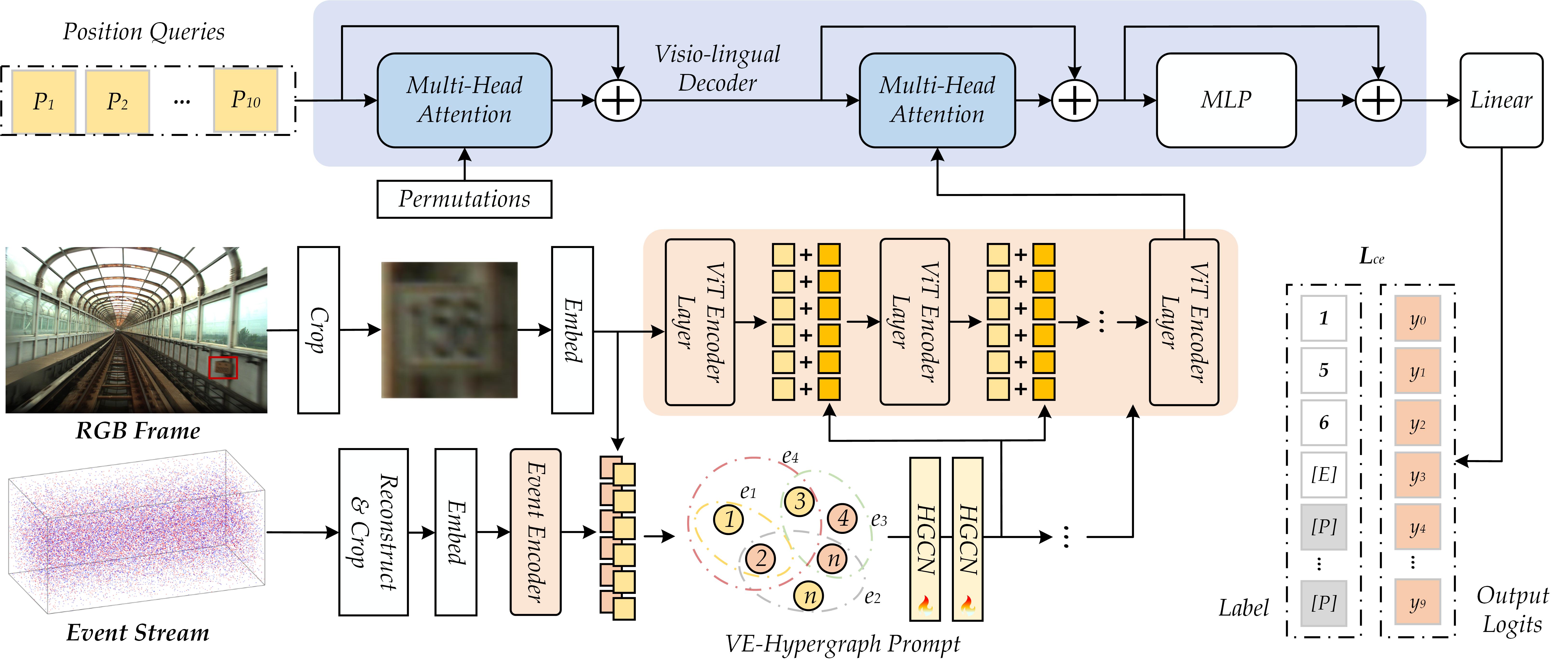}
\caption{An overview of our proposed RGB-Event based Hypergraph Prompt for Kilometer Marker Recognition based on foundation models.}
\label{framework}
\end{figure*}

\subsection{Large Foundation Models} 
Large foundation models emerge as a transformative paradigm in computer vision and multi-modal learning, as they acquire universal representations from large-scale pretraining and adapt effectively to diverse downstream tasks. ViT~\cite{dosovitskiy2020image} and Swin Transformer~\cite{liu2021swin}, as among the earliest vision foundation models to explore self-attention-based architectures, have achieved remarkable success. 
CLIP~\cite{radford2021learning} represents a pioneering vision-language model, which trains with contrastive learning on image-text pairs to align modalities and achieves strong zero-shot recognition and retrieval performance. 
Building upon this framework, SigLIP~\cite{zhai2023sigmoid} introduces a sigmoid-based matching loss in place of the conventional softmax contrastive loss. 
BLIP-2~\cite{li2023blip} further advances vision-language modeling by adopting a lightweight query transformer that bridges frozen pretrained vision encoders and large language models, thereby reducing training costs while enhancing performance in both image understanding and text generation tasks.

Beyond contrastive and bridging architectures, generative foundation models also demonstrate remarkable capabilities. 
Stable Diffusion~\cite{rombach2022high} exemplifies this trend through a latent diffusion framework that efficiently synthesizes high-fidelity images from textual descriptions and drives substantial progress in text-to-image generation. More recently, vision-language large models such as QwenVL-2.5~\cite{wang2024qwen2} extend the scope of foundation models by unifying multi-modal pretraining across large-scale image-text data. Although these large RGB-based models have achieved significant success, they are still limited by the imaging performance of RGB cameras in challenging scenarios. The proposed multi-modal hypergraph fusion approach effectively leverages the capabilities of large foundation models while incorporating event streams to mitigate the adverse effects of such challenges.

\section{Methodology} \label{sec::method}

\subsection{Overview} 

In this section, we will give an overview of the proposed HGP-KMR framework, which is designed to effectively leverage RGB frames and event streams for accurate recognition of metro mileage. As illustrated in Fig.~\ref{framework}, we first reconstruct the event stream into grayscale images and perform pre-cropping together with the RGB frames to obtain local image regions containing mileage characters. Two embedding layers are employed to project the RGB frames and grayscale images into token sequences, which are then augmented with positional encodings and jointly fed into the backbone network. Specifically, in the event branch, the features extracted by the event encoder, together with the token embeddings from the RGB modality, are fed into a hypergraph-based prompt module for cross-modal interaction. The resulting multi-modal high-order representations are subsequently integrated into the RGB backbone in a layer-wise manner, enabling a hypergraph prompt based multi-modal fusion. Finally, the encoded features are fed into a visio-lingual decoder for decoding, and the final output is obtained through a linear projection.

\subsection{Input Representation} 
Our HGP-KMR framework takes an RGB frame and the corresponding asynchronous event stream from the event camera as input. Following~\cite{rebecq2019high}, to effectively leverage existing deep neural networks for visual information modeling, we first reconstruct the corresponding event stream into an grayscale image, denoted as $E \in \mathbb{R}^{C \times H \times W}$, and represent the corresponding RGB frame as $R \in \mathbb{R}^{C \times H' \times W'}$. Subsequently, to mitigate background interference, we perform pre-cropping on both the RGB frame and the grayscale image. The cropped images are then resized to a fixed resolution (e.g., $32 \times 128$) and jointly fed into the deep network for representation learning.

\subsection{Network Architecture}

\noindent\textbf{Backbone Network}
Given the preprocessed RGB and grayscale images, denoted as $R_i \in \mathbb{R}^{3 \times h \times w}$ and $E_i \in \mathbb{R}^{3 \times h \times w}$, respectively, we first partition the images into patches of equal size according to a predefined patch size (e.g., $4 \times 8$). These patches are then projected into discrete token sequences via embedding layers and augmented with positional encodings to preserve the location information of patches, denoted as $T_r \in \mathbb{R}^{N \times C'}$ and $T_e \in \mathbb{R}^{N \times C'}$. This facilitates their input into the backbone network for cross-patch feature learning and interaction. Subsequently, the embedded features of the event modality are fed into the event encoder, which employs $L$ stacked ViT blocks as its backbone, to extract the initial event representations $F_e \in \mathbb{R}^{N \times C'}$. The formula of ViT is as follows (Layer-Normalization operation is omitted in the equation):
\begin{equation}
\begin{aligned}
         \label{vit}
         {X}_{l}^{\prime} &= \mathrm{MHA}({X}_{l-1}) + {X}_{l-1}, \quad (l = 1, 2, ... , L) \\
         {X}_{l} &= \mathrm{FFN}({X}_l^{\prime}) + {X}_l^{\prime} \\
         &= \mathrm{FFN}(\mathrm{MHA}({X}_{l-1}) + {X}_{l-1}) + \mathrm{MHA}({X}_{l-1}) + {X}_{l-1},
\end{aligned}
\end{equation}
where MHA and FFN denote the attention and feed-forward networks, respectively. $L$ is the total number of ViT blocks, ${X}_{l}$ denote the output from $l$-th Transformer block.

To enable effective interaction between the RGB and event modalities, we leverage a hypergraph network to fuse the embedded RGB features with the encoded event representations, allowing the model to capture complex cross-modal relationships. For the RGB branch, we employ a shared backbone consisting of $L$ stacked ViT blocks to extract RGB features. Meanwhile, the multi-modal information from the hypergraph network is progressively integrated into each ViT block via the hypergraph prompt strategy, ensuring that the RGB feature representations are continuously enriched with event-derived cues throughout the hypergraph network. This design allows for both modality-specific feature extraction and effective cross-modal fusion, thereby enhancing the representation capacity of the multi-modal recognition framework.

\noindent\textbf{Hypergraph Prompt}
The advantage of event cameras in extreme scenarios can significantly improve the accuracy of character recognition. Therefore, we propose a novel hypergraph-based fusion prompt module, which not only enables effective integration of multi-modal information but also enhances the feature representation capability of the RGB backbone. To obtain a unified multi-modal representation, we concatenate the RGB embedding features with the grayscale image features generated by the event encoder along the channel dimension and obtain $T_{re}$. Following~\cite{chami2019hyperbolic}, we also adopt a hypergraph construction method based on Euclidean distance K-nearest neighbors (K-NN, where the maximum number of neighboring nodes is set to 10 in our work) to generate the hypergraph structure $\mathbf{G}$. The formula can be shown as:

\begin{equation}
\begin{aligned}
\mathcal{H} &= \{V, \mathcal{E}, \mathbf{H}\}, \\[4pt]
\mathbf{H}_{j i} &= 
\begin{cases}
1, & \text{if } v_j \in e_i, \\
0, & \text{otherwise},
\end{cases} \\[6pt]
\mathbf{G} &= \mathbf{D}_v^{-\tfrac{1}{2}} \, \mathbf{H} \, \mathbf{D}_e^{-1} \, \mathbf{H}^\top \, \mathbf{D}_v^{-\tfrac{1}{2}},
\end{aligned}
\end{equation}
where the $\mathcal{H}$ refer to the hypergraph, $V$ and $\mathcal{E}$ denote the vertex and hyperedge set. $\mathbf{H}$ is the node–hyperedge incidence matrix, $\mathbf{D}_v$ and $\mathbf{D}_e$ are the node degree matrix and hyperedge degree matrix, respectively. $\mathbf{G}$ denotes the normalized hypergraph propagation matrix.

Subsequently, we feed $\mathbf{G}$ into a two-layer hypergraph convolutional neural network~\cite {chami2019hyperbolic} (HGCN) to aggregate features, which can be expressed by the following formula:

\begin{equation}
\begin{aligned}
\mathbf{X}' = \mathbf{G} (\mathbf{X} \mathbf{W} + \mathbf{b}),
\end{aligned}
\end{equation}
where the $\mathbf{X}$ and $\mathbf{X}'$ denote the input and output node feature vector, i.e., $T_{re}$ and $T_{re}'$, respectively. $\mathbf{W}$ is the learnable weight matrix and $\mathbf{b}$ is the bias. $\mathbf{G}$ is the normalized hypergraph propagation matrix.

After processing through two layers of HGCNs, we obtain the aggregated multi-modal graph features, denoted as $T_{re}'$. To further enhance the learning of the RGB modality, these features are injected into the ViT backbone of the RGB branch. This integration allows the backbone to leverage the rich multi-modal context, effectively guiding the RGB feature extraction process. Specifically, we adopt a layer-wise residual addition strategy. In this approach, for each ViT block, the RGB features produced by the block are treated as the baseline, to which the hypergraph-aggregated multi-modal features are added. The resulting fused features are then propagated to the subsequent blocks. By applying this procedure iteratively across all layers, the RGB features are progressively informed by the multi-modal context, from the shallow to the deeper layers. This hierarchical, layer-wise fusion not only encourages the backbone to capture complementary information from different modalities but also significantly enhances the final feature representation capability, leading to more discriminative embeddings for downstream tasks.

\noindent\textbf{Vision-Lingual Decoder}  
After obtaining the multi-modal image features, following~\cite{bautista2022PARseq}, we similarly employ a pre-LayerNorm Transformer decoder~\cite{baevski2018adaptive, wang2019learning} network with twice the number of attention heads, i.e., $nhead = d_{model} / 32$. The image features, position queries, and input context (i.e., \underline{1} \underline{5} \underline{6}) are fed simultaneously into the decoder. 

Following~\cite{bautista2022PARseq}, different permutations are employed as an extension of autoregressive (AR) language modeling. By applying attention masks corresponding to various permutations of the context sequence, the Transformer can learn multiple sequence factorizations during training. Unlike standard AR models, which generate sequences in a fixed order, this approach trains the model on a subset of randomly sampled permutations, enabling each position to conditionally depend on different contexts during prediction. This not only enhances the model’s understanding of sequence structures and generalization ability but also, through the masking mechanism, allows for efficient parallel computation. Position queries encode the specific target positions to be predicted, with each token directly corresponding to a particular output position, thereby enabling the model to fully exploit the benefits of different permutations.

As shown in the upper part of Fig.~\ref{framework}, let $T$ be the length of the context, the permuted context, and the position queries are jointly fed into the first attention module. The formula can be written as:
\begin{equation}
\begin{aligned}
h_c = p + \mathrm{MHA}(p, c, c, m) \in \mathbb{R}^{(T+1) \times d_{\text{model}}}. 
\end{aligned}
\end{equation}
The positional queries are denoted by $p \in \mathbb{R}^{(T+1) \times d_{\text{model}}}$, while $c \in \mathbb{R}^{(T+1) \times d_{\text{model}}}$ represents the context embeddings that include positional information. An optional attention mask is given by $m \in \mathbb{R}^{(T+1) \times (T+1)}$. Adding special delimiter tokens, such as $[B]$ or $[E]$, increases the total sequence length to $T+1$.

The second attention module is used to aggregate the features of the positional queries and the image tokens:
\begin{equation}
\begin{aligned}
h_i = h_c + \mathrm{MHA}(h_c, z, z) \in \mathbb{R}^{(T+1) \times d_{\text{model}}},
\end{aligned}
\end{equation}
where $z$ denotes the image tokens. $h_i$ is the last decoder hidden state, and is then further fed into an $\mathrm{MLP}$:
\begin{equation}
\begin{aligned}
h_{dec} = h_i + \mathrm{MLP}(h_i) \in \mathbb{R}^{(T+1) \times d_{\text{model}}}.
\end{aligned}
\end{equation}

Finally, the $h_{dec}$ through a $\mathrm{Linear}$ layer to generate the output logits:
\begin{equation}
\begin{aligned}
y = \mathrm{Linear}(h_{dec}) \in \mathbb{R}^{(T+1) \times (C+1)},
\end{aligned}
\end{equation}
where $C$ is the size of the charset.

\noindent \textbf{Loss Function} 
The Cross-Entropy loss is employed to constrain the predictions to align with the ground truth as closely as possible, which is formulated as:
\begin{equation}
\mathcal{L}_{\text{CE}} = - \sum_{i=1}^{N} \sum_{c=1}^{C} y_{i,c} \log \hat{y}_{i,c},
\end{equation}
where $N$ is the number of samples, $C$ is the size of the charset, $y_{i,c}$ is the one-hot ground truth label, and $\hat{y}_{i,c}$ is the predicted probability for class $c$.

\begin{figure*}[t]
    \centering
    \includegraphics[width=0.95\textwidth]{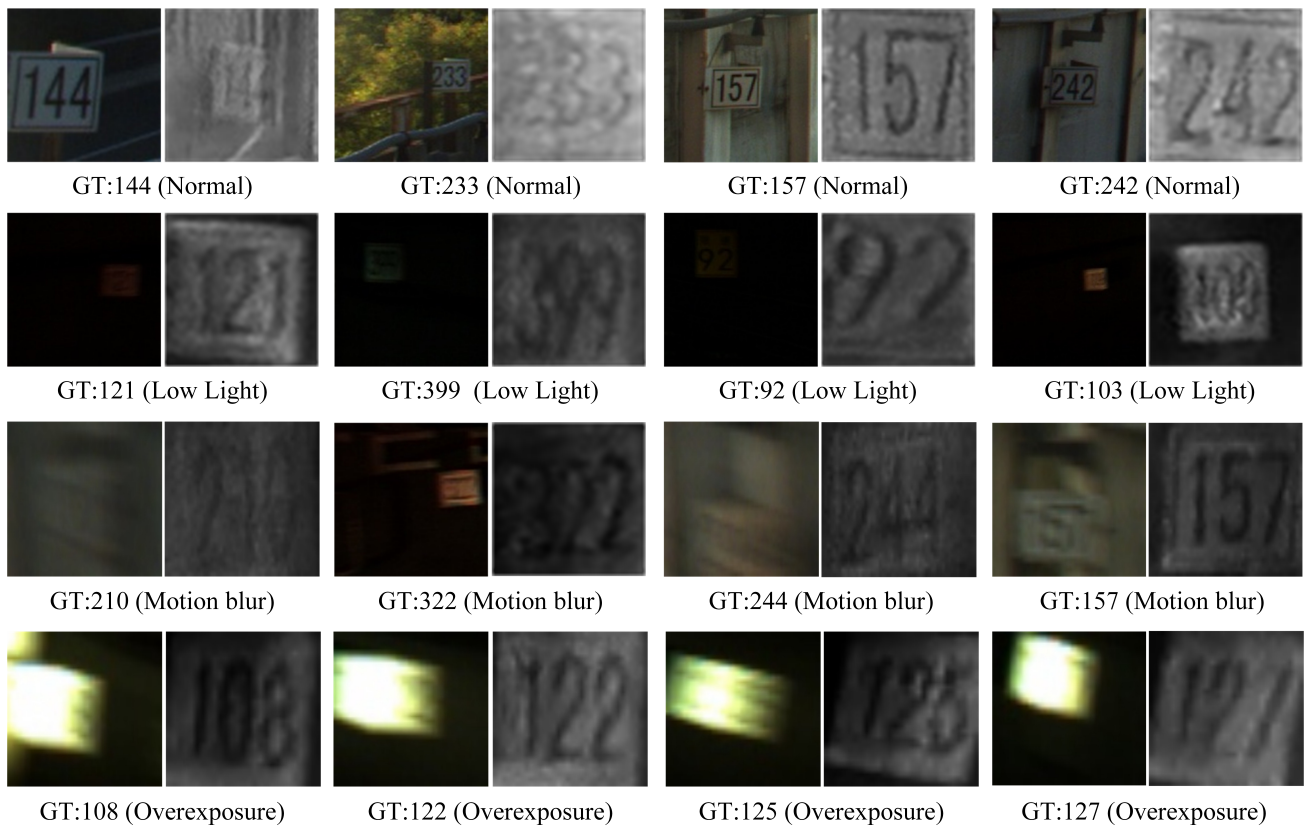}
    \caption{Example samples from the EvMetro5K dataset. 
    Each pair shows the RGB image (left) and the corresponding event-reconstructed grayscale image (right). 
    While the RGB modality often suffers from low light, motion blur, and overexposure conditions, 
    the event modality provides clearer structural information for milestone recognition.}
    \label{fig:dataset}
\end{figure*}

\section{Benchmark Dataset} \label{sec::benchmark}

\subsection{Data Collection and Annotation}  
The \textbf{EvMetro5K} dataset is collected in real metro operation scenarios using both a standard RGB camera and an event-based camera mounted in parallel. A total of 67,602 frames are recorded. The event streams are first reconstructed into grayscale images, after which all frames are manually annotated to identify visible milestones. 
The samples without milestone information are removed. To further emphasize the regions of interest, the remaining frames are cropped around the annotated milestones.

After this processing pipeline, the dataset contains \textbf{5,599} paired RGB and event-reconstructed grayscale images. 
This final dataset forms the basis for the subsequent training and evaluation of milestone recognition models.

\subsection{Statistical Analysis} 
The 5,599 samples in EvMetro5K are split into training and testing subsets at an 8:2 ratio, resulting in \textbf{4,479} training samples and \textbf{1,120} testing samples.  The dataset covers diverse metro environments with variations in illumination, motion speed, and background complexity, reflecting real-world challenges for milestone recognition.  Fig.~\ref{fig:dataset} presents representative RGB-Event pairs from the dataset. While RGB images are generally clear under normal daylight, they often suffer from motion blur during high-speed operation or visibility degradation in low-light conditions. In contrast, event-reconstructed images preserve sharper structural details, providing complementary information that supports robust milestone recognition.

\begin{table*}
\centering 
\small 
\caption{The accuracy comparisons with SOTA methods on the EvMetro5K dataset.}
\label{EvMetro5K}
\begin{tabular}{c|c|c|c|c|c}
\hline 
\textbf{Algorithm}  & \textbf{Publish}  & \textbf{Backbone}  & \textbf{Accuracy} & \textbf{Params} &\textbf{Code} \\
\hline 
IPAD~\cite{IPAD2025}  & IJCV2025  & ViT  & 69.5 &  14.1M &\href{https://github.com/Xiaomeng-Yang/IPAD}{URL} \\
Qwen3VL~\cite{Qwen2.5-VL}  & arXiv2025  & ViT  & 39.1 &  235.0B &\href{https://github.com/QwenLM/Qwen3-VL}{URL} \\
CAM~\cite{yang2024class}  & PR2024  & ResNet+ViT  & 81.3 &  23.3M &\href{https://github.com/MelosY/CAM }{URL} \\
CCD~\cite{Guan2023CCD}  & ICCV 2023  & ViT  & 86.0 &  32.0M&\href{https://github.com/TongkunGuan/CCD}{URL} \\
SIGA~\cite{guan2023SIGA}  & CVPR 2023  & ResNet& 81.3 &  40.4M&\href{https://github.com/TongkunGuan/SIGA}{URL} \\
CDistNet~\cite{zheng2024cdistnet}  & IJCV 2023  & ResNet+ViT  &  91.9  &65.5M
&\href{https://github.com/simplify23/CDistNet?tab=readme-ov-file}{URL} \\
DiG~\cite{yang2022DiG}  & ACM MM 2022  & ViT  &84.0  &  52.0M&\href{https://github.com/ayumiymk/DiG}{URL} \\
MGP-STR~\cite{wang2022mgpstr}& ECCV 2022& ViT & 92.3 &148.0M
&\href{https://github.com/AlibabaResearch/AdvancedLiterateMachinery/tree/main/OCR/MGP-STR}{URL}\\
\hline 
PARSeq~\cite{bautista2022PARseq}  & ECCV 2022  & ViT  &91.7  &  23.4M&\href{https://github.com/baudm/parseq}{URL} \\
Ours &  -& ViT  &95.1   &24.2M    & - \\ 
\hline 
\end{tabular}
\end{table*}

\begin{table*}
\centering 
\small 
\caption{The accuracy comparisons with SOTA methods on WordArt* and IC15*.}
\label{WordArt_IC15}
\begin{tabular}{c|c|c|c|c|c|c}
\hline 
\multirow{2}{*}{\textbf{Algorithm}}  & \multirow{2}{*}{\textbf{Publish}}  & \multirow{2}{*}{\textbf{Backbone}}  & \multicolumn{2}{c|}{\textbf{Accuracy}} &  \multirow{2}{*}{\textbf{Params(M)}}&\multirow{2}{*}{\textbf{Code}} \\
\cline{4-5}  
&  &  & \textbf{WordArt*}& \textbf{IC15*}&   &\\
\hline 
CCD~\cite{Guan2023CCD}  & ICCV 2023  & ViT  &62.1  & 91.6 &  52.0&\href{https://github.com/TongkunGuan/CCD}{URL} \\
SIGA~\cite{guan2023SIGA}  & CVPR 2023  & ResNet &70.9  &73.7  &  40.4&\href{https://github.com/TongkunGuan/SIGA}{URL} \\
DiG~\cite{yang2022DiG}  & ACM MM 2022  & ViT  &74.1  &78.2  &  52.0&\href{https://github.com/ayumiymk/DiG}{URL} \\
MGP-STR~\cite{wang2022mgpstr}& ECCV 2022& ViT& 80.5& 84.9& 148.0&\href{https://github.com/AlibabaResearch/AdvancedLiterateMachinery/tree/main/OCR/MGP-STR}{URL}\\
BLIVA~\cite{hu2024bliva} &AAAI 2024 &ViT &56.7 &51.3 &7531.3 & \href{https://github.com/simplify23/CDistNet?tab=readme-ov-file}{URL} \\
SimC-ESTR~\cite{wang2025eventstr} &arXiv 2025 &ViT &65.1 &56.8 &7531.3 &\href{https://github.com/Event-AHU/EventSTR}{URL} \\
ESTR-CoT~\cite{wang2025estrCoT} &arXiv 2025 &ViT &65.6 &57.1 &7531.3 &\href{https://github.com/Event-AHU/ESTR-CoT}{URL} \\  
\hline 
PARSeq~\cite{bautista2022PARseq}  & ECCV 2022  & ViT  & 74.8  &81.2  &  23.4&\href{https://github.com/baudm/parseq}{URL} \\
Ours (PARSeq) &  -& ViT  &77.8 &84.8   &24.2   &- \\ 
\hline
CDistNet~\cite{zheng2024cdistnet}  & IJCV 2023  & ResNet+ViT  &  90.8&  91.2&  65.5&\href{https://github.com/simplify23/CDistNet?tab=readme-ov-file}{URL} \\ 
Ours (CDistNet) &  -& ResNet+ViT  &91.5 &92.9   &66.2   &-  \\ 
\hline 
\end{tabular}
\end{table*}

\subsection{Benchmark Baselines}  

To provide a comprehensive evaluation on the EvMetro5K dataset, we compare our approach with a set of representative state-of-the-art (SOTA) scene text recognition models.  
These baselines are selected to cover diverse architectural paradigms, ranging from CNN-based models to transformer- and ViT-based architectures, ensuring a fair and balanced comparison.  
Specifically, we include CCD~\cite{Guan2023CCD}, SIGA~\cite{guan2023SIGA}, CDistNet~\cite{zheng2024cdistnet}, DiG~\cite{yang2022DiG}, PARSeq~\cite{bautista2022PARseq}, and MGP-STR~\cite{wang2022mgpstr}.  
These methods represent different design choices, such as self-supervised segmentation (CCD, SIGA), character-level positional modeling (CDistNet), sequence modeling via ViT and transformers (DiG, PARSeq, MGP-STR).  

For a fair comparison in the event-driven metro scenario, we extend all baseline models into a unified dual-modality framework that integrates both RGB and event modalities.  
In this way, each baseline benefits from complementary visual cues while preserving its original architectural characteristics.  
We report accuracy on EvMetro5K along with backbone architecture, parameter size, and code availability, as summarized in Table~\ref{EvMetro5K}.  
This benchmark provides a clear view of the performance-efficiency trade-offs among existing STR models under dual-modality settings and highlights the robustness challenges posed by event-driven metro scene text recognition.

\section{Experiments} \label{sec::experiments}

\subsection{Datasets and Evaluation Metric}  
In this paper, our experiments are conducted on the \textbf{WordArt*}~\cite{xie2022toward}, \textbf{IC15*}~\cite{karatzas2015icdar}, and our newly proposed \textbf{EvMetro5K} dataset. 
WordArt* is converted from the original WordArt~\cite{xie2022CornerTransformer} dataset using the event simulator ESIM~\cite{rebecq2018esim}. This dataset includes artistic text samples from posters, greeting cards, covers, and handwritten notes. It contains 4,805 training images and 1,511 validation images. 
IC15* is derived from the ICDAR2015~\cite{karatzas2015icdar} dataset and converted to event-based images. It includes 4,468 training samples and 2,077 test samples from natural scenes. 
The accuracy is used for evaluating our proposed model and other state-of-the-art recognition algorithms.

\subsection{Implementation Details}  
Our framework is built upon the PARseq architecture~\cite{bautista2022PARseq}. Based on this, we developed a novel multi-modal RGB-Event Kilometer Marker Recognition framework, which significantly improves the accuracy of mileage recognition. Our code is implemented using Python based on the PyTorch~\cite{paszke2019pytorch} framework. Adam~\cite{kingma2014adam} optimizer is used together with the 1cycle~\cite{smith2019super} learning rate scheduler, and the learning rate is set to $7e-4$. The input images are uniformly resized to $32 \times 128$ and divided into patches of size $4 \times 8$. The number of permutations is set to 6 in this work. 
The experiments are conducted on a server with a CPU Intel(R) Xeon(R) Gold 5318Y CPU @2.10GHz and GPU RTX3090s.

\begin{table*}
\centering
\small
\begin{minipage}{0.32\textwidth}
    \centering
    \caption{Ablation studies of different modalities.} 
    \label{modalities_ablation}
    \resizebox{\columnwidth}{!}{ 
    \begin{tabular}{l|c} 
    \hline 
    \textbf{\# Modalities}  &\textbf{Accuracy}    \\
    \hline
    \text{1. RGB only}  &84.2  \\
    \text{2. Event only}                &84.6  \\
    \text{3. RGB \& Event image}                &86.7  \\
    \textbf{\text{4. RGB \& Grayscale image}}              &\textbf{95.1} \\
    \hline
    \end{tabular}
    }
\end{minipage} \hfill
\begin{minipage}{0.25\textwidth}
    \centering
    \caption{Ablation studies of different GNNs.} 
    \label{fusion_layers}
    \resizebox{\columnwidth}{!}{ 
    \begin{tabular}{l|c}
    \toprule
    \textbf{Fusion Layers} & \textbf{Accuracy} \\
    \hline
    \text{1. GraphSAGE}  & 93.2  \\
    \text{2. GATConv}  & 93.7 \\
    \textbf{3. HGCN}  & \textbf{95.1} \\
    \hline
    \end{tabular}
    }
\end{minipage} \hfill
\begin{minipage}{0.35\textwidth}
    \centering
    \caption{{Ablation studies of different fusion methods.}}
    \label{fusion_methods}
    \resizebox{\columnwidth}{!}{ 
    \begin{tabular}{l|c|c} 
    \hline 
    \textbf{\# Fusion Methods}  &\textbf{Accuracy}  &\textbf{FPS}  \\
    \hline
    \text{1. Addition}    &91.7  &352  \\
    \text{2. Concatenate}    &93.5  &234  \\
    \text{3. HyperGraph Fusion}    &93.3  &97  \\
    \textbf{\text{4. HyperGraph Prompt}}   &\textbf{95.1} &89 \\
    \hline
    \end{tabular}
    }
\end{minipage} \hfill
\end{table*}

\begin{figure*}
\centering
\includegraphics[width=\textwidth]{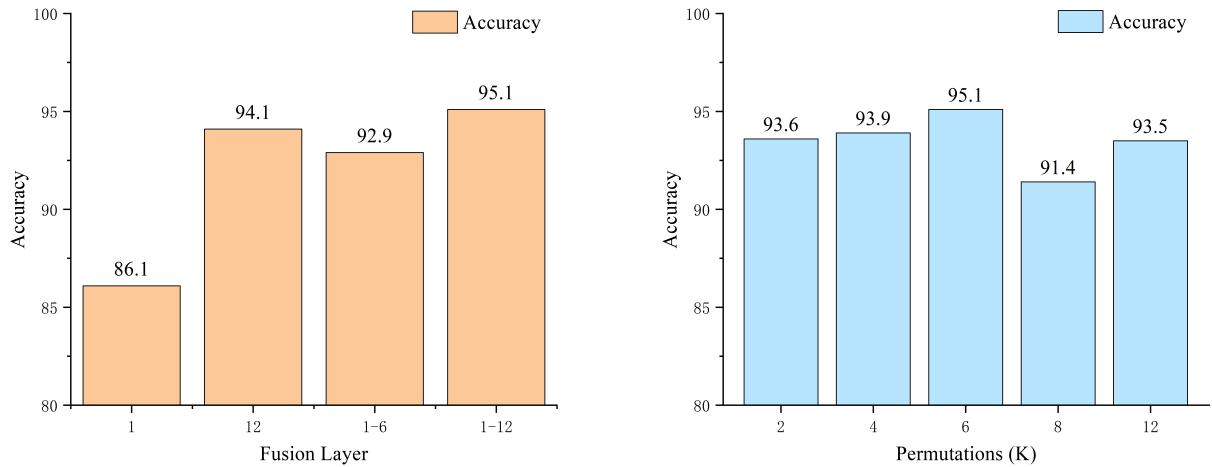}
\caption{Ablation studies of the Fusion layers and Permutations.
}
\label{Fusion layers and Permutations}
\end{figure*}

\begin{figure*}[t]
    \centering
    \includegraphics[width=\textwidth]{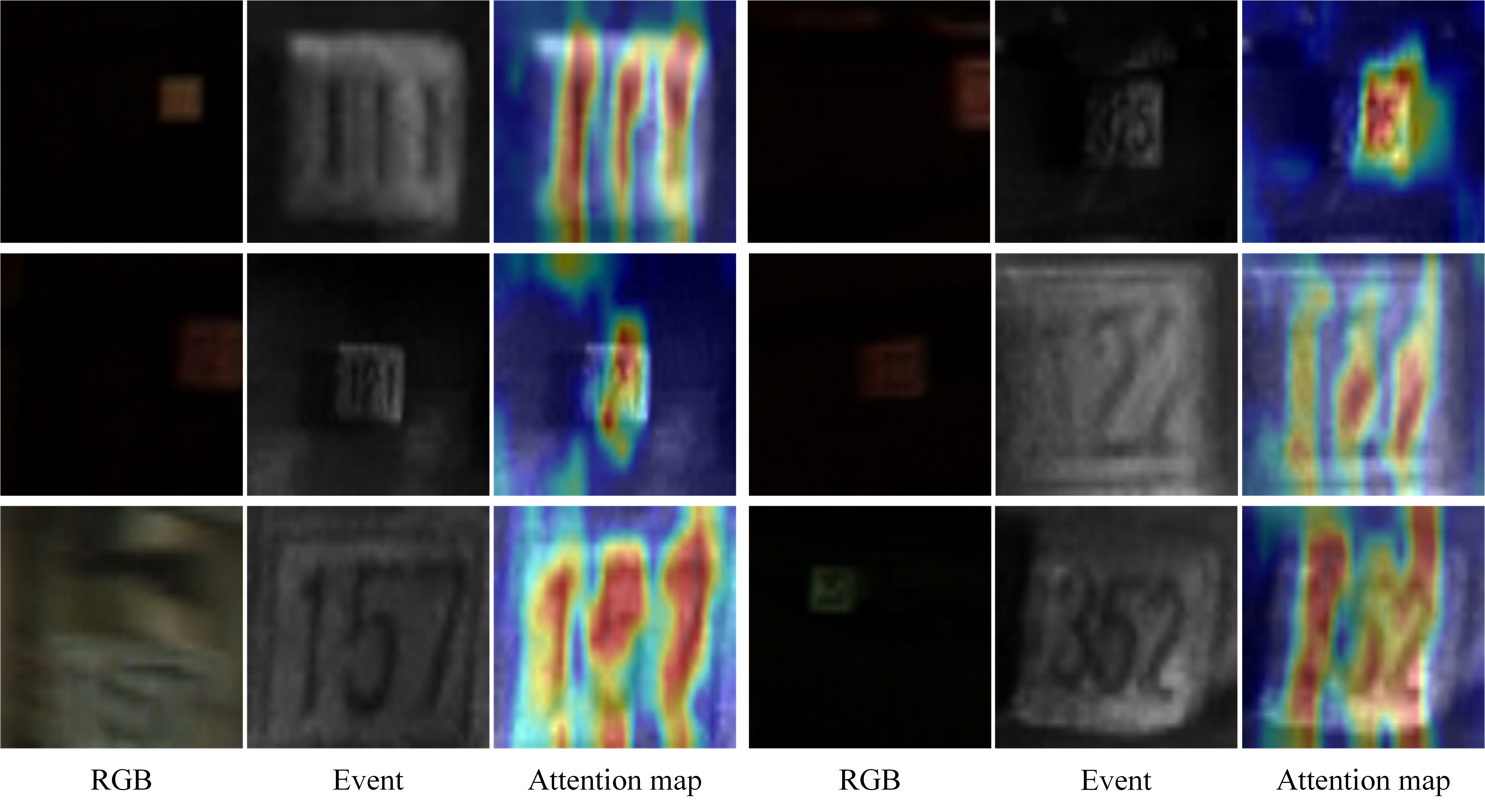}
    \caption{Visualization of attention maps of our method on the EvMetro5K dataset.}
    \label{fig:attn_maps}
\end{figure*}

\subsection{Comparison on Public Benchmarks} 

\noindent\textbf{Results on our proposed EvMetro5K dataset.}
As shown in Tab.~\ref{EvMetro5K}, we also report the recognition results of our HGP-KMR method with other SOTA recognition algorithms. It is evident that our method outperforms all compared algorithms, achieving an accuracy of 95.1\%. Compared to the baseline method, ParSeq~\cite{bautista2022PARseq}, our approach improves accuracy by +3.4\%. These results demonstrate that in challenging scenarios such as low-light conditions and fast motion, where RGB cameras typically struggle, our method effectively integrates event information, leading to a significant improvement for Kilometer Marker Recognition.

\noindent\textbf{Results on WordArt* dataset.}
This dataset is designed for artistic character recognition, with images collected from a variety of scenes, including posters, greeting cards, covers, and handwritten text. The diverse and challenging nature of these scenes adds to the difficulty of the dataset. As shown in the Tab.~\ref{WordArt_IC15}, CDistNet~\cite{zheng2024cdistnet} equipped with our hypergraph prompt strategy outperforms other methods on this dataset, achieving an accuracy of 91.5\%. Compared to other character recognition methods, such as MGP-STR and SIGA, our approach exceeds their accuracy by +11\% and +20.6\%, respectively. In addition, building upon ParSeq~\cite{bautista2022PARseq}, incorporating our prompt strategy also resulted in a significant improvement. These results demonstrate the effectiveness of our method in tackling the challenging task of artistic character recognition.

\noindent\textbf{Results on IC15* dataset.}
This dataset is a widely used benchmark for text detection and recognition tasks. It contains a large number of image samples with text information from various complex scenes, making it suitable for training and evaluating text detection and recognition algorithms. As shown in the Tab.~\ref{WordArt_IC15}, on this dataset, CDistNet~\cite{zheng2024cdistnet} equipped with our hypergraph prompt strategy significantly outperforms other state-of-the-art text recognition methods, achieving an accuracy of 92.9\%, which demonstrates the robustness of our approach across diverse scenes. Furthermore, building upon ParSeq~\cite{bautista2022PARseq}, the incorporation of our fusion strategy yields an additional +3.6\% improvement.

\subsection{Ablation Study}


\noindent\textbf{Analysis of different modalities.}
As shown in Tab.~\ref{modalities_ablation}, we compared the model’s performance across different input modalities. Here, RGB only and Event only indicate that only a single modality is input, following the training procedure of PARseq~\cite{bautista2022PARseq}. Their accuracies reached 84.2\% and 84.6\%, respectively. RGB \& Event image denote that we input the RGB image along with the event image, which is stacked according to the event stream. In our framework, the RGB frame and the reconstructed grayscale image from the event stream are fed into the network together, providing richer and more comprehensive information for mileage recognition. As a result, the recognition accuracy is significantly improved, reaching 95.1\%.

\noindent\textbf{Analysis of different GNNs.}
In this work, we employed a hypergraph to facilitate multi-modal information interaction and leveraged graph neural networks (GNNs) for learning. Different GNN architectures yield varying benefits. Specifically, we compared two common GNNs, GraphSAGE~\cite{hamilton2017inductive} and GATConv~\cite{velickovic2017graph}, and the experiments show that both achieved strong performance, with accuracies of 93.2\% and 93.7\%, respectively. To enable more effective hypergraph modeling, we ultimately adopted HGCN~\cite{feng2019hypergraph} as the graph learning network, which demonstrated the best performance.

\noindent\textbf{Analysis of fusion methods.}
We also analyzed different fusion strategies in an effort to identify the most effective method for multi-modal information interaction. First, we experimented with common approaches such as directly adding or concatenating the two modality features after extraction, which did not yield satisfactory results. We then explored concatenating the features along the channel dimension after both modalities passed through their respective encoders, followed by hypergraph neural network–based relational modeling. Finally, in our proposed framework, we performed hypergraph modeling on the RGB embeddings and the event features obtained from the encoder, and progressively integrated them into the RGB backbone branch via a prompt mechanism. As reported in the Tab.~\ref{fusion_methods}, our proposed hypergraph prompt strategy achieved the best performance. {In addition, we compare the inference speed of the model under different fusion strategies. The results show that, compared to simple addition or concatenation, the proposed multi-modal fusion method introduces a modest computational overhead, resulting in an overall inference speed of 89 FPS.}

\noindent\textbf{Analysis of fusion layers.}
We further observed that, within the hypergraph prompt strategy, integrating features into different ViT layers of the RGB branch encoder leads to varying effects. As shown in the left part of Fig.~\ref{Fusion layers and Permutations}, integrating features in the early layers (e.g., from the 1st to the 6th layer) generally results in suboptimal performance. In contrast, integrating the prompt at the final layer (the 12th layer) yields a clear performance improvement. Ultimately, the results demonstrate that integrating multi-modal hypergraph features at each layer of the RGB backbone can achieve the best performance.

\noindent\textbf{Analysis of the number of permutations.}
In the decoder part, we adopted the permutation strategy similar to PARseq~\cite{bautista2022PARseq}. We illustrated in the right part of Fig.~\ref{Fusion layers and Permutations} how different values of $K$, the number of permutations, affect the final experimental results in our work. It can be observed that the best performance is achieved when $K$ is set to 6. We argue that a smaller $K$ reflects fewer possible context orderings, which may lead to underfitting, while a larger $K$ may cause the model to overfit, resulting in suboptimal performance.

\subsection{Parameter Analysis} 
Our method also exhibits a clear advantage in terms of parameter efficiency. As reported in Tab.~\ref{EvMetro5K}, our proposed approach contains 24.2 M parameters. Building upon the ParSeq algorithm, the introduction of our hypergraph prompt strategy increases the parameter count by only 0.8 M, yet yields a substantial improvement (+3.4\%) in recognition accuracy. Furthermore, compared with other recognition methods such as MGP-STR and CDistNet, our approach uses merely 16\% and 37\% of their parameters, respectively. Collectively, these analyses indicate that the proposed framework attains superior performance under a compact parameter budget, thereby substantiating its effectiveness and efficiency.

\subsection{Visualization} 
In addition to the aforementioned analytical experiments, we also conducted visualization experiments to help readers better understand our task. As shown in the Fig.~\ref{fig:attn_maps}, we visualized the attention activation maps of the RGB encoder and overlaid them on the grayscale images. The red-highlighted regions denote the areas of focus attended to by the model. From the visualization, it can be seen that our method is able to attend to the target metro mileage character regions in various scenarios, demonstrating the robustness of our approach.

\subsection{Limitation Analysis}  
Although our method has demonstrated excellent performance across various datasets, there is still room for improvement. {On the one hand, the proposed method does not incorporate scenario-specific adaptive adjustments for different conditions. For example, under overexposure scenarios, it does not explicitly employ adaptive event filtering or intensity-aware modulation mechanisms to mitigate the loss of critical information.} On the other hand, we have not leveraged the powerful capabilities of large multi-modal models for joint modeling of visual and textual modalities. In future work, we plan to make our model adaptive to different challenging scenarios and integrate large models to achieve more accurate Kilometer Marker Recognition.

\section{Conclusion}  \label{sec::conclusion}
In this work, we have made a comprehensive contribution to the RGB-Event based Kilometer Marker Recognition (KMR) task. We introduce EvMetro5K, the first multi-modal RGB–Event mileage recognition dataset comprising 5,599 synchronized RGB–Grayscale image pairs, establishing a reliable benchmark for future research. Building upon this dataset, we propose the HGP-KMR method, a new multi-modal fusion strategy that performs hypergraph modeling on RGB and event modalities and integrates the resulting graph-structured features into the RGB backbone for layer-wise fusion. Extensive experiments across multiple datasets demonstrate the superior performance of our approach, offering a solid foundation for future developments in multi-modal mileage recognition.

\section*{Acknowledgment} 
This work was supported by the National Key R\&D Program of China (Grant No. 2022YFC3803700), the National Natural Science Foundation of China (Grant No. 62432002), and Anhui Provincial Natural Science Foundation - Outstanding Youth Project (2408085Y032). The authors acknowledge the High-performance Computing Platform of Anhui University for providing computing resources.

\small{ 
\bibliographystyle{IEEEtran}
\bibliography{reference}

\begin{thebibliography}{10}
\providecommand{\url}[1]{#1}
\csname url@samestyle\endcsname
\providecommand{\newblock}{\relax}
\providecommand{\bibinfo}[2]{#2}
\providecommand{\BIBentrySTDinterwordspacing}{\spaceskip=0pt\relax}
\providecommand{\BIBentryALTinterwordstretchfactor}{4}
\providecommand{\BIBentryALTinterwordspacing}{\spaceskip=\fontdimen2\font plus
\BIBentryALTinterwordstretchfactor\fontdimen3\font minus
  \fontdimen4\font\relax}
\providecommand{\BIBforeignlanguage}[2]{{%
\expandafter\ifx\csname l@#1\endcsname\relax
\typeout{** WARNING: IEEEtran.bst: No hyphenation pattern has been}%
\typeout{** loaded for the language `#1'. Using the pattern for}%
\typeout{** the default language instead.}%
\else
\language=\csname l@#1\endcsname
\fi
#2}}
\providecommand{\BIBdecl}{\relax}
\BIBdecl

\bibitem{he2022masked}
K.~He, X.~Chen, S.~Xie, Y.~Li, P.~Doll{\'a}r, and R.~Girshick, ``Masked
  autoencoders are scalable vision learners,'' in \emph{Proceedings of the
  IEEE/CVF conference on computer vision and pattern recognition}, 2022, pp.
  16\,000--16\,009.

\bibitem{radford2021learning}
A.~Radford, J.~W. Kim, C.~Hallacy, A.~Ramesh, G.~Goh, S.~Agarwal, G.~Sastry,
  A.~Askell, P.~Mishkin, J.~Clark \emph{et~al.}, ``Learning transferable visual
  models from natural language supervision,'' in \emph{International conference
  on machine learning}.\hskip 1em plus 0.5em minus 0.4em\relax PmLR, 2021, pp.
  8748--8763.

\bibitem{dosovitskiy2020image}
A.~Dosovitskiy, ``An image is worth 16x16 words: Transformers for image
  recognition at scale,'' \emph{arXiv preprint arXiv:2010.11929}, 2020.

\bibitem{gallego2020eventsurvey}
G.~Gallego, T.~Delbr{\"u}ck, G.~Orchard, C.~Bartolozzi, B.~Taba, A.~Censi,
  S.~Leutenegger, A.~J. Davison, J.~Conradt, K.~Daniilidis \emph{et~al.},
  ``Event-based vision: A survey,'' \emph{IEEE transactions on pattern analysis
  and machine intelligence}, vol.~44, no.~1, pp. 154--180, 2020.

\bibitem{tang2022revisiting}
C.~Tang, X.~Wang, J.~Huang, B.~Jiang, L.~Zhu, J.~Zhang, Y.~Wang, and Y.~Tian,
  ``Revisiting color-event based tracking: A unified network, dataset, and
  metric,'' \emph{arXiv preprint arXiv:2211.11010}, 2022.

\bibitem{wang2024event}
X.~Wang, S.~Wang, C.~Tang, L.~Zhu, B.~Jiang, Y.~Tian, and J.~Tang, ``Event
  stream-based visual object tracking: A high-resolution benchmark dataset and
  a novel baseline,'' in \emph{Proceedings of the IEEE/CVF Conference on
  Computer Vision and Pattern Recognition}, 2024, pp. 19\,248--19\,257.

\bibitem{huang2024mamba}
J.~Huang, S.~Wang, S.~Wang, Z.~Wu, X.~Wang, and B.~Jiang, ``Mamba-fetrack:
  Frame-event tracking via state space model,'' in \emph{Chinese Conference on
  Pattern Recognition and Computer Vision (PRCV)}.\hskip 1em plus 0.5em minus
  0.4em\relax Springer, 2024, pp. 3--18.

\bibitem{tomy2022fusing}
A.~Tomy, A.~Paigwar, K.~S. Mann, A.~Renzaglia, and C.~Laugier, ``Fusing
  event-based and rgb camera for robust object detection in adverse
  conditions,'' in \emph{2022 International Conference on Robotics and
  Automation (ICRA)}.\hskip 1em plus 0.5em minus 0.4em\relax IEEE, 2022, pp.
  933--939.

\bibitem{zhou2022rgb}
Z.~Zhou, Z.~Wu, R.~Boutteau, F.~Yang, C.~Demonceaux, and D.~Ginhac, ``Rgb-event
  fusion for moving object detection in autonomous driving,'' \emph{arXiv
  preprint arXiv:2209.08323}, 2022.

\bibitem{wang2024hardvs}
X.~Wang, Z.~Wu, B.~Jiang, Z.~Bao, L.~Zhu, G.~Li, Y.~Wang, and Y.~Tian,
  ``Hardvs: Revisiting human activity recognition with dynamic vision
  sensors,'' in \emph{Proceedings of the AAAI Conference on Artificial
  Intelligence}, vol.~38, no.~6, 2024, pp. 5615--5623.

\bibitem{wang2025rgb}
X.~Wang, H.~Wang, S.~Wang, Q.~Chen, J.~Jin, H.~Song, B.~Jiang, and C.~Li,
  ``Rgb-event based pedestrian attribute recognition: A benchmark dataset and
  an asymmetric rwkv fusion framework,'' \emph{arXiv preprint
  arXiv:2504.10018}, 2025.

\bibitem{wang2025sign}
X.~Wang, Y.~Li, F.~Wang, B.~Jiang, Y.~Wang, Y.~Tian, J.~Tang, and B.~Luo,
  ``Sign language translation using frame and event stream: Benchmark dataset
  and algorithms,'' \emph{arXiv preprint arXiv:2503.06484}, 2025.

\bibitem{ding2023mlb}
S.~Ding, J.~Chen, Y.~Wang, Y.~Kang, W.~Song, J.~Cheng, and Y.~Cao, ``E-mlb:
  Multilevel benchmark for event-based camera denoising,'' \emph{IEEE
  Transactions on Multimedia}, vol.~26, pp. 65--76, 2023.

\bibitem{wang2020eventsr}
L.~Wang, T.-K. Kim, and K.-J. Yoon, ``Eventsr: From asynchronous events to
  image reconstruction, restoration, and super-resolution via end-to-end
  adversarial learning,'' in \emph{Proceedings of the IEEE/CVF conference on
  computer vision and pattern recognition}, 2020, pp. 8315--8325.

\bibitem{rebecq2019high}
H.~Rebecq, R.~Ranftl, V.~Koltun, and D.~Scaramuzza, ``High speed and high
  dynamic range video with an event camera,'' \emph{IEEE transactions on
  pattern analysis and machine intelligence}, vol.~43, no.~6, pp. 1964--1980,
  2019.

\bibitem{bautista2022PARseq}
D.~Bautista and R.~Atienza, ``Scene text recognition with permuted
  autoregressive sequence models,'' in \emph{European conference on computer
  vision}.\hskip 1em plus 0.5em minus 0.4em\relax Springer, 2022, pp. 178--196.

\bibitem{wang2025evraindrop}
F.~Wang, F.~Zhang, X.~Wang, M.~Wang, D.~Huang, and J.~Tang, ``Evraindrop:
  Hypergraph-guided completion for effective frame and event stream
  aggregation,'' \emph{arXiv preprint arXiv:2511.21439}, 2025.

\bibitem{gao2024hypergraph}
Y.~Gao, J.~Lu, S.~Li, Y.~Li, and S.~Du, ``Hypergraph-based multi-view action
  recognition using event cameras,'' \emph{IEEE Transactions on Pattern
  Analysis and Machine Intelligence}, vol.~46, no.~10, pp. 6610--6622, 2024.

\bibitem{wang2011end}
K.~Wang, B.~Babenko, and S.~Belongie, ``End-to-end scene text recognition,'' in
  \emph{2011 International conference on computer vision}.\hskip 1em plus 0.5em
  minus 0.4em\relax IEEE, 2011, pp. 1457--1464.

\bibitem{shi2016end}
B.~Shi, X.~Bai, and C.~Yao, ``An end-to-end trainable neural network for
  image-based sequence recognition and its application to scene text
  recognition,'' \emph{IEEE transactions on pattern analysis and machine
  intelligence}, vol.~39, no.~11, pp. 2298--2304, 2016.

\bibitem{liao2019scene}
M.~Liao, J.~Zhang, Z.~Wan, F.~Xie, J.~Liang, P.~Lyu, C.~Yao, and X.~Bai,
  ``Scene text recognition from two-dimensional perspective,'' in
  \emph{Proceedings of the AAAI conference on artificial intelligence},
  vol.~33, no.~01, 2019, pp. 8714--8721.

\bibitem{han2024spotlight}
X.~Han, J.~Gao, C.~Yang, Y.~Yuan, and Q.~Wang, ``Spotlight text detector:
  Spotlight on candidate regions like a camera,'' \emph{IEEE Transactions on
  Multimedia}, 2024.

\bibitem{zhao2024E2STR}
Z.~Zhao, J.~Tang, C.~Lin, B.~Wu, C.~Huang, H.~Liu, X.~Tan, Z.~Zhang, and
  Y.~Xie, ``Multi-modal in-context learning makes an ego-evolving scene text
  recognizer,'' in \emph{Proceedings of the IEEE/CVF Conference on Computer
  Vision and Pattern Recognition}, 2024, pp. 15\,567--15\,576.

\bibitem{Guan2023CCD}
T.~Guan, W.~Shen, X.~Yang, Q.~Feng, Z.~Jiang, and X.~Yang, ``Self-supervised
  character-to-character distillation for text recognition,'' in \emph{2023
  IEEE/CVF International Conference on Computer Vision (ICCV)}, 2023, pp.
  19\,416--19\,427.

\bibitem{guan2023SIGA}
T.~Guan, C.~Gu, J.~Tu, X.~Yang, Q.~Feng, Y.~Zhao, and W.~Shen,
  ``Self-supervised implicit glyph attention for text recognition,'' in
  \emph{Proceedings of the IEEE/CVF Conference on Computer Vision and Pattern
  Recognition}, 2023, pp. 15\,285--15\,294.

\bibitem{zheng2024cdistnet}
T.~Zheng, Z.~Chen, S.~Fang, H.~Xie, and Y.-G. Jiang, ``Cdistnet: Perceiving
  multi-domain character distance for robust text recognition,''
  \emph{International Journal of Computer Vision}, vol. 132, no.~2, pp.
  300--318, 2024.

\bibitem{li2024volter}
J.-N. Li, X.-Q. Liu, X.~Luo, and X.-S. Xu, ``Volter: Visual collaboration and
  dual-stream fusion for scene text recognition,'' \emph{IEEE Transactions on
  Multimedia}, 2024.

\bibitem{Wei2024BUSNet}
J.~Wei, H.~Zhan, Y.~Lu, X.~Tu, B.~Yin, C.~Liu, and U.~Pal, ``Image as a
  language: Revisiting scene text recognition via balanced, unified and
  synchronized vision-language reasoning network,'' in \emph{Proceedings of the
  AAAI Conference on Artificial Intelligence}, vol.~38, no.~6, 2024, pp.
  5885--5893.

\bibitem{na2022matrn}
B.~Na, Y.~Kim, and S.~Park, ``Multi-modal text recognition networks:
  Interactive enhancements between visual and semantic features,'' in
  \emph{European Conference on Computer Vision}.\hskip 1em plus 0.5em minus
  0.4em\relax Springer, 2022, pp. 446--463.

\bibitem{da2022levocr}
C.~Da, P.~Wang, and C.~Yao, ``Levenshtein ocr,'' in \emph{European Conference
  on Computer Vision}.\hskip 1em plus 0.5em minus 0.4em\relax Springer, 2022,
  pp. 322--338.

\bibitem{fang2021ABINet}
S.~Fang, H.~Xie, Y.~Wang, Z.~Mao, and Y.~Zhang, ``Read like humans: Autonomous,
  bidirectional and iterative language modeling for scene text recognition,''
  in \emph{Proceedings of the IEEE/CVF conference on computer vision and
  pattern recognition}, 2021, pp. 7098--7107.

\bibitem{liu2024textmonkey}
Y.~Liu, B.~Yang, Q.~Liu, Z.~Li, Z.~Ma, S.~Zhang, and X.~Bai, ``Textmonkey: An
  ocr-free large multimodal model for understanding document,'' \emph{arXiv
  preprint arXiv:2403.04473}, 2024.

\bibitem{feng2023docpedia}
H.~Feng, Q.~Liu, H.~Liu, W.~Zhou, H.~Li, and C.~Huang, ``Docpedia: Unleashing
  the power of large multimodal model in the frequency domain for versatile
  document understanding,'' \emph{arXiv preprint arXiv:2311.11810}, 2023.

\bibitem{wei2025vary}
H.~Wei, L.~Kong, J.~Chen, L.~Zhao, Z.~Ge, J.~Yang, J.~Sun, C.~Han, and
  X.~Zhang, ``Vary: Scaling up the vision vocabulary for large vision-language
  model,'' in \emph{European Conference on Computer Vision}.\hskip 1em plus
  0.5em minus 0.4em\relax Springer, 2025, pp. 408--424.

\bibitem{hu2024mplug}
A.~Hu, H.~Xu, J.~Ye, M.~Yan, L.~Zhang, B.~Zhang, C.~Li, J.~Zhang, Q.~Jin,
  F.~Huang \emph{et~al.}, ``mplug-docowl 1.5: Unified structure learning for
  ocr-free document understanding,'' \emph{arXiv preprint arXiv:2403.12895},
  2024.

\bibitem{wei2024OCR2.0}
H.~Wei, C.~Liu, J.~Chen, J.~Wang, L.~Kong, Y.~Xu, Z.~Ge, L.~Zhao, J.~Sun,
  Y.~Peng \emph{et~al.}, ``General ocr theory: Towards ocr-2.0 via a unified
  end-to-end model,'' \emph{arXiv preprint arXiv:2409.01704}, 2024.

\bibitem{wang2025eventstr}
X.~Wang, J.~Jiang, D.~Li, F.~Wang, L.~Zhu, Y.~Wang, Y.~Tian, and J.~Tang,
  ``Eventstr: A benchmark dataset and baselines for event stream based scene
  text recognition,'' \emph{arXiv preprint arXiv:2502.09020}, 2025.

\bibitem{wang2025estrCoT}
X.~Wang, J.~Jiang, Q.~Chen, L.~Chen, L.~Zhu, Y.~Wang, Y.~Tian, and J.~Tang,
  ``Estr-cot: Towards explainable and accurate event stream based scene text
  recognition with chain-of-thought reasoning,'' \emph{arXiv preprint
  arXiv:2507.02200}, 2025.

\bibitem{zhang2022data}
D.~Zhang, Q.~Ding, P.~Duan, C.~Zhou, and B.~Shi, ``Data association between
  event streams and intensity frames under diverse baselines,'' in
  \emph{European Conference on Computer Vision}.\hskip 1em plus 0.5em minus
  0.4em\relax Springer, 2022, pp. 72--90.

\bibitem{chen2025evhdr}
Z.~Chen, Z.~Liao, D.~Ma, H.~Tang, Q.~Zheng, and G.~Pan, ``Evhdr-nerf: Building
  high dynamic range radiance fields with single exposure images and events,''
  in \emph{Proceedings of the AAAI Conference on Artificial Intelligence},
  vol.~39, no.~3, 2025, pp. 2376--2384.

\bibitem{yang2023learning}
Y.~Yang, J.~Han, J.~Liang, I.~Sato, and B.~Shi, ``Learning event guided high
  dynamic range video reconstruction,'' in \emph{Proceedings of the IEEE/CVF
  Conference on Computer Vision and Pattern Recognition}, 2023, pp.
  13\,924--13\,934.

\bibitem{chen2024event}
Z.~Chen, Z.~Lu, D.~Ma, H.~Tang, X.~Jiang, Q.~Zheng, and G.~Pan, ``Event-id:
  Intrinsic decomposition using an event camera,'' in \emph{Proceedings of the
  32nd ACM International Conference on Multimedia}, 2024, pp. 10\,095--10\,104.

\bibitem{liang2024towards}
G.~Liang, K.~Chen, H.~Li, Y.~Lu, and L.~Wang, ``Towards robust event-guided
  low-light image enhancement: a large-scale real-world event-image dataset and
  novel approach,'' in \emph{Proceedings of the IEEE/CVF Conference on Computer
  Vision and Pattern Recognition}, 2024, pp. 23--33.

\bibitem{lu2025edygs}
M.~Lu, Z.~Chen, Y.~Liu, D.~Ma, H.~Tang, Q.~Zheng, and G.~Pan, ``Edygs: Event
  enhanced dynamic 3d radiance fields from blurry monocular video,'' in
  \emph{Proceedings of the Thirty-Fourth International Joint Conference on
  Artificial Intelligence}, 2025, pp. 1684--1692.

\bibitem{liu2025nemf}
Y.~Liu, Z.~Chen, H.~Yan, D.~Ma, H.~Tang, Q.~Zheng, and G.~Pan, ``E-nemf:
  Event-based neural motion field for novel space-time view synthesis of
  dynamic scenes,'' in \emph{Proceedings of the IEEE/CVF International
  Conference on Computer Vision}, 2025, pp. 10\,854--10\,864.

\bibitem{yan2025evstvsr}
H.~Yan, Z.~Lu, Z.~Chen, D.~Ma, H.~Tang, Q.~Zheng, and G.~Pan, ``Evstvsr: Event
  guided space-time video super-resolution,'' in \emph{Proceedings of the AAAI
  Conference on Artificial Intelligence}, vol.~39, no.~9, 2025, pp. 9085--9093.

\bibitem{ma2024color4e}
Y.~Ma, P.~Duan, Y.~Hong, C.~Zhou, Y.~Zhang, J.~Ren, and B.~Shi, ``Color4e:
  Event demosaicing for full-color event guided image deblurring,'' in
  \emph{Proceedings of the 32nd ACM International Conference on Multimedia},
  2024, pp. 661--670.

\bibitem{li2024coarse}
H.~Li, H.~Shi, and X.~Gao, ``A coarse-to-fine fusion network for event-based
  image deblurring,'' in \emph{Proceedings of the International Joint
  Conference on Artificial Intelligence}, 2024, pp. 974--982.

\bibitem{guo2025utnet}
F.~Guo, P.~Ren, and C.~Luo, ``Utnet: event-rgb multimodal fusion model for
  underwater transparent organism detection,'' \emph{Intelligent Marine
  Technology and Systems}, vol.~3, no.~1, p.~18, 2025.

\bibitem{qi2023e2nerf}
Y.~Qi, L.~Zhu, Y.~Zhang, and J.~Li, ``E2nerf: Event enhanced neural radiance
  fields from blurry images,'' in \emph{Proceedings of the IEEE/CVF
  International Conference on Computer Vision}, 2023, pp. 13\,254--13\,264.

\bibitem{fan2025efficient}
L.~Fan, J.~Yang, L.~Wang, J.~Zhang, X.~Lian, and H.~Shen, ``Efficient spiking
  neural network for rgb--event fusion-based object detection,''
  \emph{Electronics}, vol.~14, no.~6, p. 1105, 2025.

\bibitem{liu2021swin}
Z.~Liu, Y.~Lin, Y.~Cao, H.~Hu, Y.~Wei, Z.~Zhang, S.~Lin, and B.~Guo, ``Swin
  transformer: Hierarchical vision transformer using shifted windows,'' in
  \emph{Proceedings of the IEEE/CVF international conference on computer
  vision}, 2021, pp. 10\,012--10\,022.

\bibitem{zhai2023sigmoid}
X.~Zhai, B.~Mustafa, A.~Kolesnikov, and L.~Beyer, ``Sigmoid loss for language
  image pre-training,'' in \emph{Proceedings of the IEEE/CVF international
  conference on computer vision}, 2023, pp. 11\,975--11\,986.

\bibitem{li2023blip}
J.~Li, D.~Li, S.~Savarese, and S.~Hoi, ``Blip-2: Bootstrapping language-image
  pre-training with frozen image encoders and large language models,'' in
  \emph{International conference on machine learning}.\hskip 1em plus 0.5em
  minus 0.4em\relax PMLR, 2023, pp. 19\,730--19\,742.

\bibitem{rombach2022high}
R.~Rombach, A.~Blattmann, D.~Lorenz, P.~Esser, and B.~Ommer, ``High-resolution
  image synthesis with latent diffusion models,'' in \emph{Proceedings of the
  IEEE/CVF conference on computer vision and pattern recognition}, 2022, pp.
  10\,684--10\,695.

\bibitem{wang2024qwen2}
P.~Wang, S.~Bai, S.~Tan, S.~Wang, Z.~Fan, J.~Bai, K.~Chen, X.~Liu, J.~Wang,
  W.~Ge \emph{et~al.}, ``Qwen2-vl: Enhancing vision-language model's perception
  of the world at any resolution,'' \emph{arXiv preprint arXiv:2409.12191},
  2024.

\bibitem{chami2019hyperbolic}
I.~Chami, R.~Ying, C.~R{\'e}, and J.~Leskovec, ``Hyperbolic graph convolutional
  neural networks,'' \emph{Advances in neural information processing systems},
  vol.~32, p. 4869, 2019.

\bibitem{baevski2018adaptive}
A.~Baevski and M.~Auli, ``Adaptive input representations for neural language
  modeling,'' \emph{arXiv preprint arXiv:1809.10853}, 2018.

\bibitem{wang2019learning}
Q.~Wang, B.~Li, T.~Xiao, J.~Zhu, C.~Li, D.~F. Wong, and L.~S. Chao, ``Learning
  deep transformer models for machine translation,'' in \emph{Proceedings of
  the 57th Annual Meeting of the Association for Computational Linguistics},
  2019, pp. 1810--1822.

\bibitem{IPAD2025}
X.~Yang, Z.~Qiao, and Y.~Zhou, ``Ipad: Iterative, parallel, and diffusion-based
  network for scene text recognition,'' \emph{International Journal of Computer
  Vision}, 2025.

\bibitem{Qwen2.5-VL}
S.~Bai, K.~Chen, X.~Liu, J.~Wang, W.~Ge, S.~Song, K.~Dang, P.~Wang, S.~Wang,
  J.~Tang, H.~Zhong, Y.~Zhu, M.~Yang, Z.~Li, J.~Wan, P.~Wang, W.~Ding, Z.~Fu,
  Y.~Xu, J.~Ye, X.~Zhang, T.~Xie, Z.~Cheng, H.~Zhang, Z.~Yang, H.~Xu, and
  J.~Lin, ``Qwen2.5-vl technical report,'' \emph{arXiv preprint
  arXiv:2502.13923}, 2025.

\bibitem{yang2024class}
M.~Yang, B.~Yang, M.~Liao, Y.~Zhu, and X.~Bai, ``Class-aware mask-guided
  feature refinement for scene text recognition,'' \emph{Pattern Recognition},
  vol. 149, p. 110244, 2024.

\bibitem{yang2022DiG}
M.~Yang, M.~Liao, P.~Lu, J.~Wang, S.~Zhu, H.~Luo, Q.~Tian, and X.~Bai,
  ``Reading and writing: Discriminative and generative modeling for
  self-supervised text recognition,'' in \emph{Proceedings of the 30th ACM
  International Conference on Multimedia}, 2022, pp. 4214--4223.

\bibitem{wang2022mgpstr}
P.~Wang, C.~Da, and C.~Yao, ``Multi-granularity prediction for scene text
  recognition,'' in \emph{European Conference on Computer Vision}.\hskip 1em
  plus 0.5em minus 0.4em\relax Springer, 2022, pp. 339--355.

\bibitem{hu2024bliva}
W.~Hu, Y.~Xu, Y.~Li, W.~Li, Z.~Chen, and Z.~Tu, ``Bliva: A simple multimodal
  llm for better handling of text-rich visual questions,'' in \emph{Proceedings
  of the AAAI Conference on Artificial Intelligence}, vol.~38, no.~3, 2024, pp.
  2256--2264.

\bibitem{xie2022toward}
X.~Xie, L.~Fu, Z.~Zhang, Z.~Wang, and X.~Bai, ``Toward understanding wordart:
  Corner-guided transformer for scene text recognition,'' in \emph{European
  conference on computer vision}.\hskip 1em plus 0.5em minus 0.4em\relax
  Springer, 2022, pp. 303--321.

\bibitem{karatzas2015icdar}
D.~Karatzas, L.~Gomez-Bigorda, A.~Nicolaou, S.~Ghosh, A.~Bagdanov, M.~Iwamura,
  J.~Matas, L.~Neumann, V.~R. Chandrasekhar, S.~Lu \emph{et~al.}, ``Icdar 2015
  competition on robust reading,'' in \emph{2015 13th international conference
  on document analysis and recognition (ICDAR)}.\hskip 1em plus 0.5em minus
  0.4em\relax IEEE, 2015, pp. 1156--1160.

\bibitem{xie2022CornerTransformer}
X.~Xie, L.~Fu, Z.~Zhang, Z.~Wang, and X.~Bai, ``Toward understanding wordart:
  Corner-guided transformer for scene text recognition,'' in \emph{European
  conference on computer vision}.\hskip 1em plus 0.5em minus 0.4em\relax
  Springer, 2022, pp. 303--321.

\bibitem{rebecq2018esim}
H.~Rebecq, D.~Gehrig, and D.~Scaramuzza, ``Esim: an open event camera
  simulator,'' in \emph{Conference on robot learning}.\hskip 1em plus 0.5em
  minus 0.4em\relax PMLR, 2018, pp. 969--982.

\bibitem{paszke2019pytorch}
A.~Paszke, S.~Gross, F.~Massa, A.~Lerer, J.~Bradbury, G.~Chanan, T.~Killeen,
  Z.~Lin, N.~Gimelshein, L.~Antiga \emph{et~al.}, ``Pytorch: An imperative
  style, high-performance deep learning library,'' \emph{Advances in neural
  information processing systems}, vol.~32, pp. 8026--8037, 2019.

\bibitem{kingma2014adam}
D.~P. Kingma and J.~Ba, ``Adam: A method for stochastic optimization,''
  \emph{arXiv preprint arXiv:1412.6980}, 2014.

\bibitem{smith2019super}
L.~N. Smith and N.~Topin, ``Super-convergence: Very fast training of neural
  networks using large learning rates,'' in \emph{Artificial intelligence and
  machine learning for multi-domain operations applications}, vol. 11006.\hskip
  1em plus 0.5em minus 0.4em\relax SPIE, 2019, pp. 369--386.

\bibitem{hamilton2017inductive}
W.~Hamilton, Z.~Ying, and J.~Leskovec, ``Inductive representation learning on
  large graphs,'' \emph{Advances in neural information processing systems},
  vol.~30, 2017.

\bibitem{velickovic2017graph}
P.~Velickovic, G.~Cucurull, A.~Casanova, A.~Romero, P.~Lio, Y.~Bengio
  \emph{et~al.}, ``Graph attention networks,'' \emph{stat}, vol. 1050, no.~20,
  pp. 10--48\,550, 2017.

\bibitem{feng2019hypergraph}
Y.~Feng, H.~You, Z.~Zhang, R.~Ji, and Y.~Gao, ``Hypergraph neural networks,''
  in \emph{Proceedings of the AAAI conference on artificial intelligence},
  vol.~33, no.~01, 2019, pp. 3558--3565.

\end{thebibliography}
}

\end{document}